\documentclass[letterpaper, 10 pt, conference]{ieeeconf} 

\IEEEoverridecommandlockouts        

\usepackage{cite}

\usepackage{amsmath,amssymb,amsfonts}
\usepackage{graphicx}
\usepackage{textcomp}
\usepackage{xcolor}
\usepackage{color}													
\usepackage{xfrac}	
\usepackage{colortbl}
\usepackage{siunitx}
\usepackage[english]{babel}	
\usepackage{multirow}
\usepackage{hyperref}
\usepackage{blindtext}
\usepackage{mathtools}
\usepackage{hyperref}
\usepackage{tablefootnote}
\usepackage{algorithmic} 
\usepackage{textcomp} 
\usepackage{gensymb} 
\usepackage[ruled, linesnumbered]{algorithm2e}
\usepackage[caption=false, font=footnotesize]{subfig}

\newcommand{\Res}[0]{\boldsymbol{\mathit{\delta}}}
\newcommand{\mm}[1]{\boldsymbol{#1}}
\newcommand{\ind}[1]{\mathrm{#1}}

\definecolor{gruen}{RGB}{0,152,70}
\definecolor{blau}{RGB}{0,0,255}
\definecolor{orange}{RGB}{255,110,0}
\definecolor{olivengruen}{RGB}{110, 117, 14}
\definecolor{gruenmarker}{RGB}{0, 158, 34}

\title{\LARGE \bf
	Safe Collision and Clamping Reaction for\\Parallel Robots During Human-Robot Collaboration
}

\author{Aran Mohammad, Moritz Schappler, Tim-Lukas Habich and Tobias Ortmaier
	\thanks{All authors are with the Leibniz University Hannover, Institute of Mechatronic Systems, 30823 Garbsen, Germany,
		{\tt\small aran.mohammad@imes.uni-hannover.de}{\newline Supporting video: \url{https://youtu.be/pcIBYYhcWk4}}}%
}
\makeatletter
\newcommand{\removelatexerror}{\let\@latex@error\@gobble}
\makeatother

\newif\ifcopyright
		\copyrighttrue
\begin{document}
	
	\ifcopyright
	{\LARGE IEEE Copyright Notice}
	\newline
	\fboxrule=0.4pt \fboxsep=3pt
	
	\fbox{\begin{minipage}{1.1\linewidth}  
			Copyright (c) 2023 IEEE. Personal use of this material is permitted. For any other purposes, permission must be obtained from the IEEE by emailing pubs-permissions@ieee.org. \\
			
			Accepted to be published in: Proceedings of the 2023 IEEE/RSJ International Conference on Intelligent Robots (IROS), October 1 -- 5, 2023, Detroit, Michigan, USA.  
			
	\end{minipage}}
	\else
	\fi
	\graphicspath{{./graphics/}}
	\maketitle
	\thispagestyle{empty}
	\pagestyle{empty}
	
	\begin{abstract}
		Parallel robots (PRs) offer the potential for safe human-robot collaboration because of their low moving masses. 
		Due to the in-parallel kinematic chains, the risk of contact in the form of collisions and clamping at a chain increases. 
		Ensuring safety is investigated in this work through various contact reactions on a real planar PR.
		External forces are estimated based on proprioceptive information and a dynamics model, which allows contact detection.
		Retraction along the direction of the estimated line of action provides an instantaneous response to limit the occurring contact forces within the experiment to $\SI{70}{\newton}$ at a maximum velocity of $\SI{0.4}{\meter / \second}$.
		A reduction in the stiffness of a Cartesian impedance control is investigated as a further strategy.
		For clamping, a feedforward neural network (FNN) is trained and tested in different joint angle configurations to classify whether a collision or clamping occurs with an accuracy of $80\%$.
		A second FNN classifies the clamping kinematic chain to enable a subsequent kinematic projection of the clamping joint angle onto the rotational platform coordinates.
		In this way, a structure opening is performed in addition to the softer retraction movement.	
		The reaction strategies are compared in real-world experiments at different velocities and controller stiffnesses to demonstrate their effectiveness.
		The results show that in all collision and clamping experiments the PR terminates the contact in less than $\SI{130}{\milli \second}$.
	\end{abstract}
	
	\section{Introduction}
		Safety in physical human-robot collaboration (HRC) is ensured by limiting the \emph{kinetic energy} resulting from the contact partners' effective \emph{mass and relative speed}~\cite{InternationalOrganizationforStandardization.2016}. 
		Accordingly, lightweight serial robots reduce the collision energy in the event of contact. 
		Another approach is using a parallel robot (PR) to reduce moving masses and maintain the \emph{same energy limits at significantly higher speeds or to decrease the demanded energy for a fixed trajectory}. 
		PRs are characterized by typically base-mounted drives with kinematic chains connected to a mobile platform~\cite{Merlet.2006}.
		\begin{figure}[h!]
			\vspace{1.5mm}			
			\centering
			\includegraphics[width=1\columnwidth]{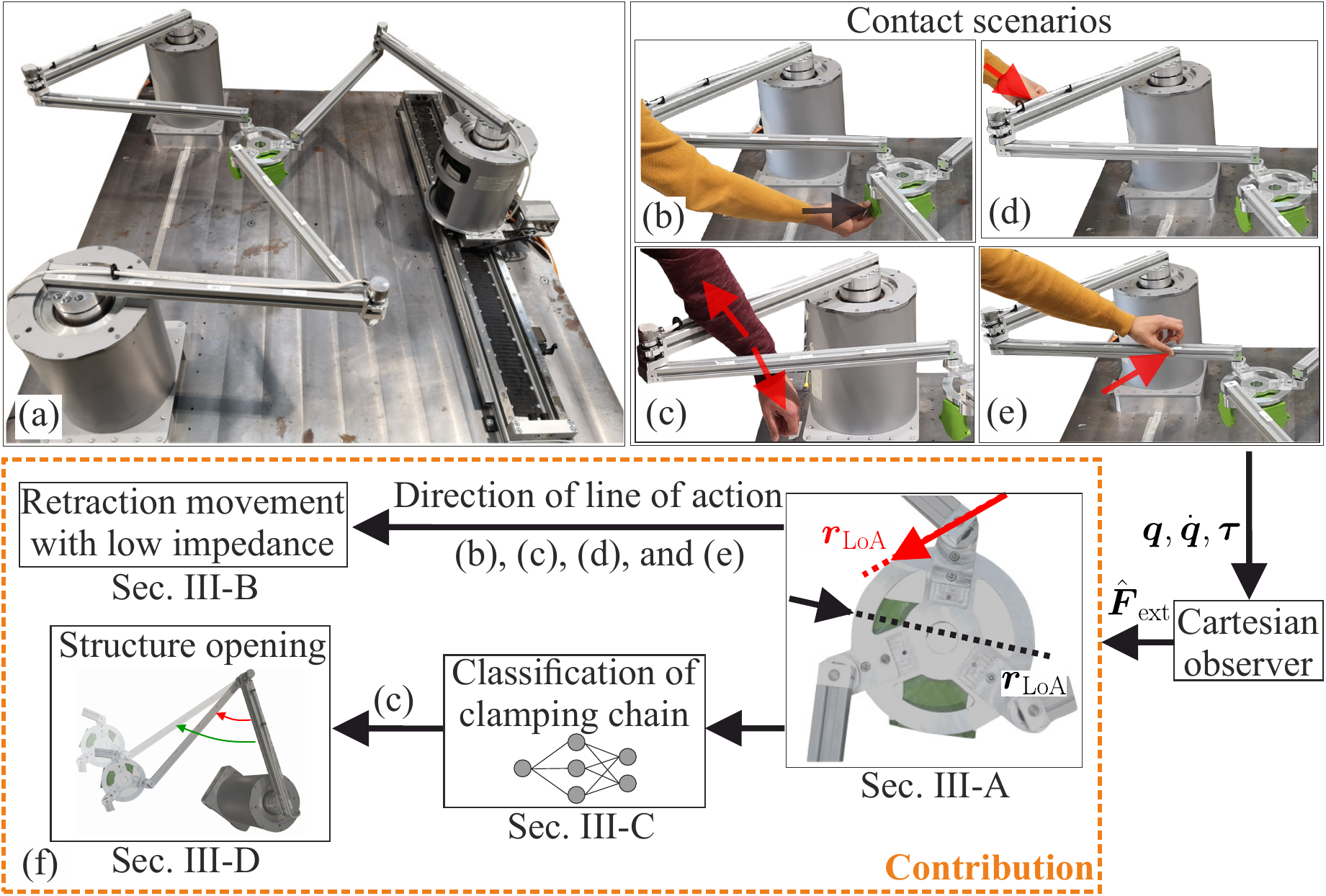}
			\caption{(a) The parallel robot considered in this work with contact scenarios: (b) Platform collision, (c) chain clamping, (d) first and (e) second link collision. (f) Contribution of this work: Based on the estimated line of action $\boldsymbol{r}_\mathrm{LoA}$, a retraction movement is realized with the translational platform coordinates and additionally low Cartesian impedances as a reaction to collisions. Two neural networks classify the clamping chain to perform a structure opening with the rotational platform coordinates.}
			\label{fig:titelbild}
			\vspace{-1.5mm}
		\end{figure}
		An example of a PR is shown in Fig.~\ref{fig:titelbild}(a).
		\subsection{Related Work}
			HRC with both serial and parallel kinematics requires \emph{contact detection} and \emph{reaction} to unwanted physical contacts. 
			These can occur as shown in Fig.~\ref{fig:titelbild} in the event of a collision across the entire structure or clamping in the leg chains. 
			The force and pressure applied to the human should comply with the thresholds specified in~\cite{InternationalOrganizationforStandardization.2016} to prevent pain, injury or death.
			This can be used as an optimization problem for robot control~\cite{Meguenani.2015, Gabrielli.2021}.
			Alternatively, the reflected mass can be minimized by reconfiguring the robot~\cite{Haddadin.2012,Mansfeld.2017}.
			
			In the case of physical contact, detection is performed with \emph{exteroceptive or proprioceptive} information.
			Tactile skin~\cite{Dahiya.2013, Albini.2017, Svarny.2022} or image-based approaches~\cite{Rosenstrauch.2018, Merckaert.2022, Hoang.2022} allow an exteroceptive contact detection. 
			Visual information combined with permissible force thresholds enables an increase in performance while maintaining safety constraints~\cite{Ferraguti.2020, Lucci.2020, Svarny.2019}.
			
			However, detection for response in \emph{dynamic contacts} must be fast and robust. 
			Therefore, proprioceptive information via built-in sensors in the robot offers advantages in terms of shorter sample times and lower hardware requirements. 			
			Physical or data-driven models enable contact detection based on the measurement of joint angles and torques. 
			Supervised machine learning algorithms can distinct intentional and accidental contacts by learning temporal and dynamic effects from time series or physically modeled features~\cite{Golz.2015, Zhang.2021}. 
			A disturbance observer of the generalized momentum provides the detection, isolation, and identification of external contacts~\cite{Luca.2003, Haddadin.2017}. 
			A physically formulated minimization problem enables the detection by comparing the external torques caused by the location and force of the contact to the measured driving torques~\cite{Likar.2014, Popov.2019, Wang.2020, Manuelli.2016}. 
			Assumptions like contact location on the link surface or no normal velocities of the collided link at the contact point constrain the optimization problem. 
			Furthermore, the temporal information in the measurements can be incorporated into a particle filter to solve ambiguous cases~\cite{Manuelli.2016, Wang.2020}.\\
			The results of detection, isolation, and identification decide on \emph{contact reactions}. 
			In~\cite{Luca.2006}, different reaction strategies are presented on a serial robot. 
			Here, a momentum observer modeled in the joint space detects and locates the collided link. 
			An admittance control then performs a reflex motion as a function of the external joint torque.
			In~\cite{Haddadin.2008}, an unintentional contact causes the trajectory to scale in time by decrementing the trajectory planning back into the past.
			Furthermore, redundancy resolution using a null-space projection can be realized to follow the reference trajectory with simultaneous admittance-controlled response~\cite{Luca.2008}.
			Once the contact location is known, a Cartesian interaction control depending on the contact force allows a reconfiguration of the robot~\cite{Magrini.2014, Magrini.2015}.
		\subsection{Contributions}
			The presented proprioceptive detection and reaction methods apply to \emph{robots with open-loop kinematics}.
			The fundamental assumptions do not apply to parallel robots, since they consist of \emph{closed kinematic chains}. 
			A contact on a kinematic chain affects multiple drives due to the coupling by the mobile platform. 
			Compared to the authors' previous work~\cite{Mohammad.2023}, where a change into the zero-g mode is considered, a retraction movement and a structure opening are contributed as new reaction methods that account for the \emph{closed-loop kinematics} of a PR. 
			As shown in Fig.~\ref{fig:titelbild}(f), a Cartesian disturbance observer is adopted to estimate the external forces' line of action leading to the contributions of this work:
		\begin{itemize}
			\item The trajectory planning and control in the operational space allow the decoupling of the translational and rotational coordinates for a retraction movement and an opening of the clamping structure. 
			\item Two feedforward neural networks for classification are designed to distinguish clamping and collision, as well as to predict the clamping kinematic chain. 
			\item A structure opening provides a reaction to clamping contact. The gradient of the critical joints' angle with respect to the platform orientation is calculated and determines the direction of the structure opening. 
			\item The methods are compared regarding different platform velocities and stiffnesses of the Cartesian impedance controller by real experiments on a planar PR.
		\end{itemize}
		The paper is organized as follows.
		Section \ref{sec:preliminaries} defines the kinematics and dynamics modeling. 
		The reaction algorithm is presented in Sec. \ref{sec:reaction}. 
		Section \ref{sec:validation} describes the PR considered in this paper, followed by an experimental evaluation of contacts on the entire structure. 
		The conclusion follows in Sec. \ref{sec:conlusions}.
	\section{Preliminaries} \label{sec:preliminaries}
		This section expresses the kinematics (\ref{ssec:kinematics}) and dynamics modeling (\ref{ssec:dynamics}) of the used PR. 
		A Cartesian impedance control (\ref{ssec:controller}) and a disturbance observer (\ref{ssec:observer}) are then described. 
		\begin{figure}[tb!]
			\vspace{1.5mm}
			\centering
			\includegraphics[width=\columnwidth]{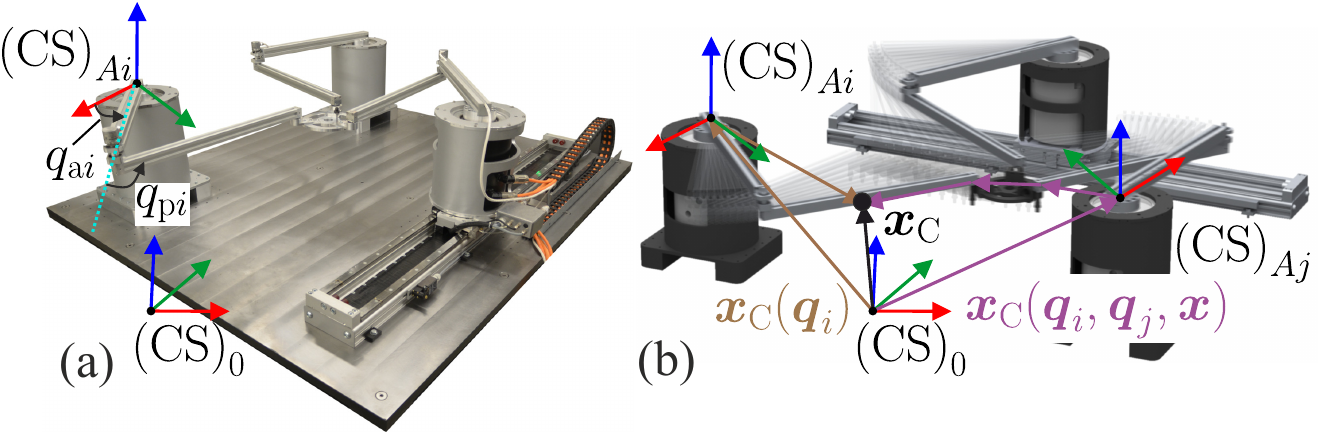}
			\caption{(a) The 3-\underline{R}RR PR from~\cite{Mohammad.2023} with (b) a contact at $\boldsymbol{x}_\mathrm{C}$ at the $i$-th leg chain --- $\boldsymbol{x}_\mathrm{C}$ can be related to any leg chain and the joint angles $\boldsymbol{q}$}
			\label{fig_3PRRR_real_skizze}
			\vspace{-1.5mm}
		\end{figure}
		\subsection{Kinematics} \label{ssec:kinematics}
			The modeling is described using the planar {3-\underline{R}RR} parallel robot\footnotemark\footnotetext{The letter R denotes a revolute joint and an underlining represents an actuated joint. The actuated prismatic joint of the parallel robot is kept constant and is therefore not considered in the modeling.}
			shown in Fig.~\ref{fig_3PRRR_real_skizze}(a) with $m{=}3$ platform degrees of freedom and $n{=}3$ leg chains\cite{Thanh.2012}. 
			The methods can be applied to any fully-parallel robot~($n{=}m$).
			Operational space coordinates (platform pose), active, passive, and coupling joint angles are respectively given by $\boldsymbol{x}^\mathrm{T}{=}[\boldsymbol{x}_\mathrm{t}^\mathrm{T}, x_\mathrm{r}]{\in}\mathbb{R}^m,\boldsymbol{q}_\mathrm{a}{\in}\mathbb{R}^n$ and $\boldsymbol{q}_\mathrm{p}, \boldsymbol{q}_\mathrm{c}{\in}\mathbb{R}^3$.
			The $n_i{=}3$ joint angles (active, passive, platform coupling) of each leg chain in $\boldsymbol{q}_i{\in}\mathbb{R}^{n_i}$ are stacked as $\boldsymbol{q}^\mathrm{T}{=}[\boldsymbol{q}_1^\mathrm{T}, \boldsymbol{q}_2^\mathrm{T}, \boldsymbol{q}_3^\mathrm{T}] {\in} \mathbb{R}^{3n}$. 
			
			By closing vector loops~\cite{Merlet.2006}, the kinematic constraints $\Res (\boldsymbol{q}, \boldsymbol{x}){=}\boldsymbol{0}$ are constructed. 
			Eliminating the passive joint angles yields the reduced kinematic constraints $\Res_\mathrm{red}(\boldsymbol{q}_\mathrm{a}, \boldsymbol{x}){=}\boldsymbol{0}$. 
			From this, the active joint angles are analytically calculated (inverse kinematics).
			Passive joint angles are measured to estimate the platform's pose and to encounter ambiguity of the forward kinematics. 
			Since the encoders' accuracies vary, the Newton-Raphson approach is then applied. 
			
			For differential kinematics, a time derivative of the kinematic constraints gives
			\begin{align}\label{eq_DifKin_Jac1}
				\dot{\boldsymbol{q}}&={-}\Res_{\partial \boldsymbol{q}}^{-1}\Res_{\partial \boldsymbol{x}} \dot{\boldsymbol{x}}=\boldsymbol{J}_{q,x}\dot{\boldsymbol{x}},\\
				\label{eq_DifKin_Jac2}
				\dot{\boldsymbol{x}}&= {-}\left(\Res_\mathrm{red}\right)_{\partial \boldsymbol{x}}^{-1} \left(\Res_\mathrm{red}\right)_{\partial \boldsymbol{q}_\mathrm{a}} \dot{\boldsymbol{q}}_\mathrm{a}=\boldsymbol{J}_{x,q_\mathrm{a}}\dot{\boldsymbol{q}}_\mathrm{a}
			\end{align}
			using the notation $\boldsymbol{a}_{\partial \boldsymbol{b}}{\coloneqq} \sfrac{\partial \boldsymbol{a}}{\partial \boldsymbol{b}}$ and the Jacobian matrices\footnotemark $\boldsymbol{J}_{q, x}{\in}\mathbb{R}^{3n\times m}$ and $\boldsymbol{J}_{x, q_\mathrm{a}}{\in}\mathbb{R}^{m\times n}$.\footnotetext{For the sake of readability, dependencies on $\boldsymbol{q}$ and $\boldsymbol{x}$ are omitted.}
			
			The kinematics of an arbitrary (contact) point on the robot structure is modeled by formulating the contact coordinates $\boldsymbol{x}_\mathrm{C}$ of a point $\mathrm{C}$ via joint angles $\boldsymbol{q}$. 
			As depicted in Fig.~\ref{fig_3PRRR_real_skizze}(b), the $i$-th chains' serial forward kinematics is obtained by $\boldsymbol{x}_\mathrm{C}{=}\boldsymbol{f}_i( \boldsymbol{q}_i)$.
			Alternatively, $\boldsymbol{x}_\mathrm{C}$ is represented via the $j$-th chain and the platform orientation by $\boldsymbol{f}_j(\boldsymbol{q}_i, \boldsymbol{q}_j, \boldsymbol{x})$.
			The latter can be substituted with the full kinematic constraints~\cite{Schappler.2019}, leading to $\boldsymbol{x}_\mathrm{C}(\boldsymbol{q}_i, \boldsymbol{q}_j)$.
			A time derivative results in $\dot{\boldsymbol{x}}_\mathrm{C}{=}\boldsymbol{J}_{x_\mathrm{C},q}\dot{\boldsymbol{q}}$ with the Jacobian $\boldsymbol{J}_{x_\mathrm{C},q}$. 
			Based on (\ref{eq_DifKin_Jac1}) and (\ref{eq_DifKin_Jac2}), the projection of the differential kinematics of platform and joint coordinates onto the contact is expressed by
			\begin{subequations}\label{eq:contactJacobian} \begin{align}
					\dot{\boldsymbol{x}}_\mathrm{C} &= \boldsymbol{J}_{x_\mathrm{C},q} \dot{ \boldsymbol{q}}\\
					&=\boldsymbol{J}_{x_\mathrm{C},q} \boldsymbol{J}_{q,x} \dot{\boldsymbol{x}} = \boldsymbol{J}_{x_\mathrm{C}, x} \dot{\boldsymbol{x}} \\ 
					&= \boldsymbol{J}_{x_\mathrm{C}, x} \boldsymbol{J}_{x, q_\mathrm{a}} \dot{\boldsymbol{q}}_\mathrm{a} = \boldsymbol{J}_{x_\mathrm{C}, q_\mathrm{a}} \dot{\boldsymbol{q}}_\mathrm{a}
			\end{align} \end{subequations}
			with the Jacobian matrices $\boldsymbol{J}_{x_\mathrm{C}, x}$ and $\boldsymbol{J}_{x_\mathrm{C}, q_\mathrm{a}}$.
		\subsection{Dynamics} \label{ssec:dynamics}
			The Lagrangian equations of the second kind, the \textit{subsystem} and \textit{coordinate partitioning} methods formulate the equations of motion in the operational space without the constraint forces~\cite{Thanh.2009}.
			The dynamics model\footnote{Generalized forces $\boldsymbol{F}{\in}\mathbb{R}^m$ (including moments) in operational space} of the PR is
			\begin{equation} \label{eq_dyn}
				\boldsymbol{M}_x \ddot{\boldsymbol{x}}{+} \boldsymbol{c}_x {+} \boldsymbol{g}_x{+} \boldsymbol{F}_{\mathrm{fr},x}= \boldsymbol{F}_\mathrm{m} {+} \boldsymbol{F}_\mathrm{ext}
			\end{equation} 
			where $\boldsymbol{M}_x$ is denoted as the inertia matrix, $\boldsymbol{c}_x{=}\boldsymbol{C}_x\dot{\boldsymbol{x}}$ as the vector/matrix of the centrifugal and Coriolis effects, $\boldsymbol{g}_x$ as the gravitational terms, $\boldsymbol{F}_{\mathrm{fr},x}$ as the viscous and Coulomb friction components, $\boldsymbol{F}_\mathrm{m}$ as the forces resulting from the motor torques and $\boldsymbol{F}_{\mathrm{ext}}$ as external forces. 
			The forces $\boldsymbol{F}_\mathrm{m}$ are projected into the actuated joint coordinates by the principle of virtual work $\boldsymbol{\tau}_\mathrm{a}{=}\boldsymbol{J}_{x,q_\mathrm{a}}^\mathrm{T}\boldsymbol{F}_\mathrm{m}$. 
			External forces $\boldsymbol{F}_\mathrm{ext,link}$ at a link affect in a configuration-dependent way the mobile platform and the actuated joints via
			\begin{subequations}\label{eq_trafo_link_mP_Drives}\begin{align} 
					\boldsymbol{F}_\mathrm{ext,mP}&=\boldsymbol{J}_{x_\mathrm{C},x}^\mathrm{T} \boldsymbol{F}_\mathrm{ext,link},\\
					\boldsymbol{\tau}_\mathrm{a,ext}&=\boldsymbol{J}_{x_\mathrm{C},q_\mathrm{a}}^\mathrm{T} \boldsymbol{F}_\mathrm{ext,link}.
			\end{align} \end{subequations}
		\subsection{Cartesian Impedance Control in Operational Space} \label{ssec:controller}
			Cartesian impedance control for PRs~\cite{Taghirad.2013} intuitively parameterizes the robot environmental dynamics in platform coordinates and is described by
			\begin{equation} \label{eq:ImpRegX}
				\boldsymbol{F}_\mathrm{m} = \hat{\boldsymbol{c}}_x + \hat{\boldsymbol{g}}_x + \hat{\boldsymbol{M}}_x \ddot{\boldsymbol{x}}_\mathrm{d} + \hat{\boldsymbol{F}}_{\mathrm{fr},x} + \boldsymbol{K}_\mathrm{d} \boldsymbol{e}_x + \boldsymbol{D}_\mathrm{d} \dot{\boldsymbol{e}}_x 
			\end{equation}
			with the compensation of the dynamics components and the pose error $\boldsymbol{e}_x{=}\boldsymbol{x}_\mathrm{d}{-}\boldsymbol{x}$ between the desired and actual pose $\boldsymbol{x}_\mathrm{d}$ and $\boldsymbol{x}$. 
			To set a specific modal damping behavior, the desired stiffness matrix ${\boldsymbol{K}_\mathrm{d}{=}\mathrm{diag}(k_{\mathrm{d},1}, \dots ,k_{\mathrm{d},m}){>}\boldsymbol{0}}$ and the inertia matrix are used for the factorization damping design~\cite{AlbuSchaffer.2003}
			\begin{align} 
				\boldsymbol{D}_\mathrm{d} = \tilde{\boldsymbol{M}}_{x} \boldsymbol{D}_\xi \tilde{\boldsymbol{K}}_\mathrm{d} + \tilde{\boldsymbol{K}}_\mathrm{d} \boldsymbol{D}_\xi \tilde{\boldsymbol{M}}_{x},
			\end{align} 
			with ${\boldsymbol{D}_\xi{=}\mathrm{diag}(D_{\xi,1}, \dots ,D_{\xi,m})}{>}\boldsymbol{0}$, $\boldsymbol{K}_\mathrm{d}{=}\tilde{\boldsymbol{K}}_\mathrm{d} \tilde{\boldsymbol{K}}_\mathrm{d}$ and $\boldsymbol{M}_x{=}\tilde{\boldsymbol{M}}_{x} \tilde{\boldsymbol{M}}_{x}$ (due to the symmetric positive-definite $\boldsymbol{M}_x$). 
			The closed-loop error dynamics results in
			\begin{align}
				{\boldsymbol{M}_x ( \ddot{\boldsymbol{x}} {-}\ddot{\boldsymbol{x}}_\mathrm{d} ) {+} \boldsymbol{D}_\mathrm{d} ( \dot{\boldsymbol{x}} {-}\dot{\boldsymbol{x}}_\mathrm{d} ) {+} \boldsymbol{K}_\mathrm{d} ( \boldsymbol{x} {-} \boldsymbol{x}_\mathrm{d} ) {=} \boldsymbol{F}_\mathrm{ext}}
			\end{align}
			with the external force as input and the pose error as output. 
		\subsection{Generalized-Momentum Observer} \label{ssec:observer}
			Introduced by~\cite{Luca.2003}, a residual of the generalized momentum $\boldsymbol{p}_x{=}\boldsymbol{M}_x \dot{\boldsymbol{x}}$ is set up in the operational space, since $\boldsymbol{x}$ represents the minimal coordinates for the dynamics of PRs. 
			The time derivative of the residual is $\sfrac{\mathrm{d}}{\mathrm{dt}} \hat{\boldsymbol{F}}_\mathrm{ext} {=} \boldsymbol{K}_{\mathrm{o}} (\dot{\boldsymbol{p}}_x {-} \dot{\hat{\boldsymbol{p}}}_x )$ with the observer gain matrix $\boldsymbol{K}_\mathrm{o}{=}\mathrm{diag}(k_{\mathrm{o},1}, \dots ,k_{\mathrm{o},m})$ and $k_{\mathrm{o},i}{>}0$. 
			The generalized-momentum observer (MO) is constructed by expressing~(\ref{eq_dyn}) as $\hat{\boldsymbol{M}}_x\ddot{\boldsymbol{x}}$ and substituting it by the term $\dot{\hat{\boldsymbol{p}}}_x$ in the time integral of $\sfrac{\mathrm{d}}{\mathrm{dt}} \hat{\boldsymbol{F}}_\mathrm{ext}$. 
			With $\dot{\hat{\boldsymbol{M}}}_x{=}\hat{\boldsymbol{C}}_x^\mathrm{T}{+}\hat{\boldsymbol{C}}_x$~\cite{Haddadin.2017, Ott.2008}, the external force estimation is realized by
			\begin{align} \label{eq:mo}
				\hat{\boldsymbol{F}}_\mathrm{ext} &= \boldsymbol{K}_\mathrm{o} \left( \hat{\boldsymbol{M}}_x \dot{\boldsymbol{x}} {-} \int_{0}^t (\boldsymbol{F}_\mathrm{m} {-} \hat{\boldsymbol{\beta}} {+} \hat{\boldsymbol{F}}_\mathrm{ext}) \mathrm{d}\tilde{t} \right), \\ 
				\label{eq:mo_beta}
				\hat{\boldsymbol{\beta}} &= \hat{\boldsymbol{g}}_x {+} \hat{\boldsymbol{F}}_{\mathrm{fr},x} {+}( \hat{\boldsymbol{C}}_x {-}\dot{\hat{\boldsymbol{M}}}_x )\dot{\boldsymbol{x}} = \hat{\boldsymbol{g}}_x {+} \hat{\boldsymbol{F}}_{\mathrm{fr},x}{-}\hat{\boldsymbol{C}}_x^\mathrm{T} \dot{\boldsymbol{x}}.
			\end{align}
			By well-identified dynamics in (\ref{eq:mo_beta}), the MO's estimation exponentially converges to the external force in the platform coordinates with the linear and decoupled error dynamics
			$\boldsymbol{K}_\mathrm{o}^{-1} \dot{\hat{\boldsymbol{F}}}_\mathrm{ext} {+} \hat{\boldsymbol{F}}_\mathrm{ext}{=}\boldsymbol{F}_\mathrm{ext}$. 
	\section{Contact Reaction} \label{sec:reaction}
		The main contributions are introduced in this section. 
		The effects of platform and link contacts on operational space coordinates are considered (\ref{ssec:Effects_CollClamp}), which provide the basis for retraction movement and low-stiffness reaction (\ref{ssec:Retraction}). 
		Next, the clamping classification is presented (\ref{ssec:ContactClass}), which initiates an opening of the clamping leg chain (\ref{ssec:StructureOpening}).
		Finally, the implementation of the proposed reaction strategy is described (\ref{ssec:Implementation}).
		\subsection{Effects of a Collision and Clamping} \label{ssec:Effects_CollClamp}
			\begin{figure}[t!]
				\vspace{1.5mm}
				\centering
				\includegraphics[width=\columnwidth]{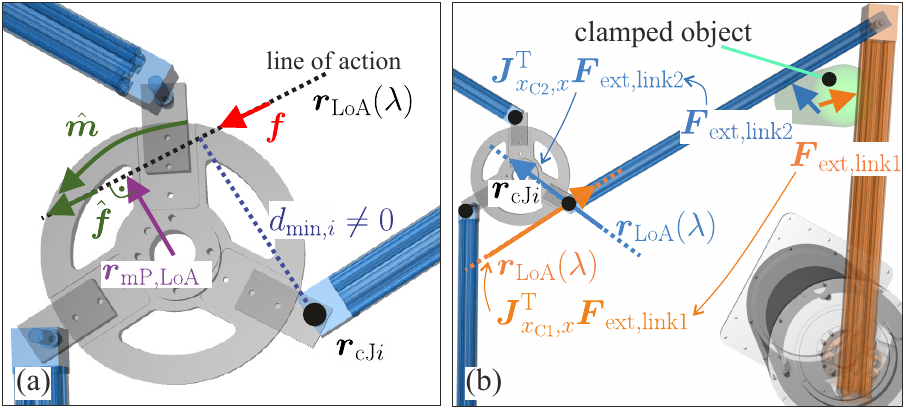}
				\caption{(a) Effects of an external force $\boldsymbol{f}$ on the platform with the estimate $\hat{\boldsymbol{F}}_\mathrm{ext}{=}(\hat{\boldsymbol{f}}^\mathrm{T}, \hat{\boldsymbol{m}}^\mathrm{T})^\mathrm{T}$, the minimum lever $\boldsymbol{r}_\mathrm{mP,LoA}$ and the line of action $\boldsymbol{r}_\mathrm{LoA}(\lambda)$ in a MuJoCo simulation~\cite{Todorov.2012}. 
				The minimum distance $d_{\min,i}$ is between $\boldsymbol{r}_\mathrm{LoA}(\lambda)$ and the coupling point $\boldsymbol{r}_{\mathrm{cJ}i}$. 
				(b) Clamped object with the link forces $\boldsymbol{F}_{\mathrm{ext,link}}$ and their projections $\boldsymbol{J}_{x_{\mathrm{C}},x}^\mathrm{T} \boldsymbol{F}_{\mathrm{ext,link}}$ on the platform coordinates}
				\label{fig:KinetostaticAnalysis}
				\vspace{-1.5mm}
			\end{figure}
			A contact at the PR with forces $\boldsymbol{f}$ and moments $\boldsymbol{m}$ is summarized as a wrench $\boldsymbol{F}_\mathrm{ext}{=}(\boldsymbol{f}^\mathrm{T}, \boldsymbol{m}^\mathrm{T})^\mathrm{T}$. 
			In a collision or clamping scenario, only forces are considered so that $\boldsymbol{m}{\equiv}\boldsymbol{0}$ holds.
			Figure~\ref{fig:KinetostaticAnalysis} shows examples of different contact scenarios and their effects on the platform coordinates. 
			While a platform contact force is expressed in the operational space coordinates (Fig.~\ref{fig:KinetostaticAnalysis}(a)), a collision force $\boldsymbol{F}_\mathrm{ext,link}$ at a link is projected onto the platform coordinates via the Jacobian matrix from (\ref{eq:contactJacobian}), see Fig~\ref{fig:KinetostaticAnalysis}(b).
			A clamping is now represented by two single link forces, which are projected to the platform coordinates with 
			\begin{equation} 
				\begin{split}
					\boldsymbol{F}_\mathrm{ext,mP}=\boldsymbol{J}_{x_{\mathrm{C}1},x}^\mathrm{T} \boldsymbol{F}_\mathrm{ext,link1} + \boldsymbol{J}_{x_{\mathrm{C}2},x}^\mathrm{T} \boldsymbol{F}_\mathrm{ext,link2}.
				\end{split}
			\end{equation}	
			Regardless of the contact scenario, the MO in (\ref{eq:mo}) estimates $\hat{\boldsymbol{F}}_\mathrm{ext}{=}(\hat{\boldsymbol{f}}^\mathrm{T}, \hat{\boldsymbol{m}}^\mathrm{T})^\mathrm{T}$ with the forces $\hat{\boldsymbol{f}}$ and moments
			\begin{equation} \label{eq:hatm_hatf}
				\begin{split}
					\hat{\boldsymbol{m}} = \boldsymbol{r} {\times} \hat{\boldsymbol{f}} = \boldsymbol{S} (\boldsymbol{r})\hat{\boldsymbol{f}} = \boldsymbol{S}^\mathrm{T} ( \hat{\boldsymbol{f}}) \boldsymbol{r},
				\end{split}
			\end{equation}
			where $\boldsymbol{S}$ is a skew-symmetric matrix, and $\boldsymbol{r}$ is a lever between the body-fixed platform coordinate system to any point on the line of action (LoA). 
			Using the Moore-Penrose inverse~$(\dagger)$ of $\boldsymbol{S}(\hat{\boldsymbol{f}})$~\cite{Haddadin.2017}, the minimum distance 
			\begin{equation} \label{eq:CalcMPLoA}
				\begin{split} 
					\boldsymbol{r}_\mathrm{mP,LoA} &= ( \boldsymbol{S}^\mathrm{T} ( \hat{\boldsymbol{f}}) )^\dagger \hat{\boldsymbol{m}}
				\end{split}
			\end{equation} 
			from the platform coordinate system to the LoA of the external force is calculated.
			The LoA 
			\begin{equation} \label{eq:CalcLoA}
				\boldsymbol{r}_\mathrm{LoA}(\lambda){=}\boldsymbol{r}_\mathrm{mP, LoA}{+}\lambda \hat{\boldsymbol{n}}_\mathrm{f}
			\end{equation} 
			can be defined with $\hat{\boldsymbol{n}}_\mathrm{f}{=}\hat{\boldsymbol{f}}/||\hat{\boldsymbol{f}} ||_2$ and the scalar variable $\lambda$.	
		\subsection{Reaction Strategy: Retraction Movement} \label{ssec:Retraction}
			As visible in~(\ref{eq:CalcLoA}), the estimate $\hat{\boldsymbol{n}}_\mathrm{f}$ contains information on the direction of the external force's LoA. 
			As a first reaction strategy, the direction $\hat{\boldsymbol{n}}_\mathrm{f}$ of the LoA leads to the calculation of a new target position 
				\begin{align} \label{eq:retr_tr}
					\tilde{\boldsymbol{x}}_\mathrm{t,d}=\boldsymbol{x}_\mathrm{t}+d_\mathrm{react} \hat{\boldsymbol{n}}_\mathrm{f}
				\end{align}
			with a predefined distance $d_\mathrm{react}$ and the current platform position $\boldsymbol{x}_\mathrm{t}$.
			In case of a collision, a trajectory planning is initiated for the new target pose $\tilde{\boldsymbol{x}}_\mathrm{d}$ and used as an input in~(\ref{eq:ImpRegX}) \emph{without requiring any information on the contact location}.
			Since the PR is controlled in the operational space and the \emph{closed-loop kinematics} are considered in the kinetostatic projection $\boldsymbol{\tau}_\mathrm{a}{=}\boldsymbol{J}_{x,q_\mathrm{a}}^\mathrm{T}\boldsymbol{F}_\mathrm{m}$, the \emph{retraction movement} occurs with all $n$ kinematic chains.	
			\begin{figure}[t!]
				\vspace{1.5mm} 
				\centering
				\includegraphics[width=\columnwidth]{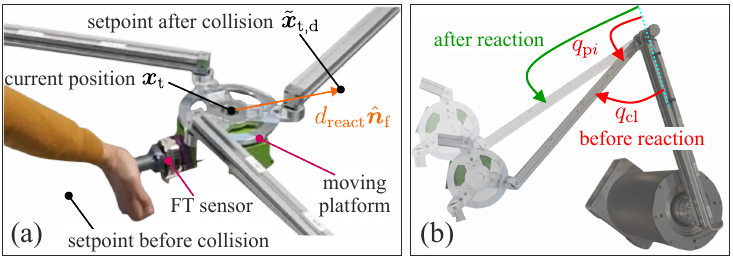}
				\caption{(a) Retraction movement (b) and structure opening after detection}
				\label{fig:reaktion_rueckz_strukoeff}
				\vspace{-1.5mm}
			\end{figure}	
			Fig.~\ref{fig:reaktion_rueckz_strukoeff}(a) shows a collision and the direction of the retraction movement along $d_\mathrm{react}\hat{\boldsymbol{n}}_\mathrm{f}$ from the platform position $\boldsymbol{x}_\mathrm{t}$.
			This reaction strategy can be performed with a \emph{reduction of the translational stiffness of the controller} $\tilde{\boldsymbol{K}}_\mathrm{t,d}{<}\boldsymbol{K}_\mathrm{t,d}$ up to $\tilde{\boldsymbol{K}}_\mathrm{t,d}{=}\boldsymbol{0}$ (zero-g mode).	
		\subsection{Classification of a Clamping Chain} \label{ssec:ContactClass}
			Two feedforward neural networks (FNNs) are trained for clamping classification using the gradient-based optimization method Adam~\cite{Kingma.22122014, FabianPedregosa.2011}. 
			The hyperbolic tangent function is chosen as the nonlinear activation function. 
			To avoid underfitting and overfitting, a hyperparameter optimization of an $L_2$ regularization term $\lambda {\ge} 0$, as well as of the network structure with the number of neurons and hidden layers is performed by a grid search.
			The FNNs are used for:
			\subsubsection{Clamping Classification}
				The first FNN classifies a contact into the output classes $\{$Collision, Clamping$\}$ based on the input data $\hat{\boldsymbol{F}}_\mathrm{ext}$ from~(\ref{eq:mo}).
				If the FNN classifies a collision, only the retraction movement is performed.
				In the case of clamping, a second FNN is initiated.
			\subsubsection{Chain Classification}					
				One difference in Fig.~\ref{fig:KinetostaticAnalysis}(b) from the platform contact in Fig.~\ref{fig:KinetostaticAnalysis}(a) is that the minimum distance 
				\begin{equation} \label{eq:MinDistd}
					\begin{split} 
						d_{\mathrm{min},i}{=}||\left(\boldsymbol{r}_{\mathrm{cJ}i}{-}\boldsymbol{r}_\mathrm{mP, LoA}\right) {\times} \hat{\boldsymbol{n}}_\mathrm{f}||_2
					\end{split}
				\end{equation} 
				from $\boldsymbol{r}_\mathrm{LoA}(\lambda)$ to the $i$-th coupling joint $\boldsymbol{r}_{\mathrm{cJ}i}$ is zero.
				The minimum distance is determined for each leg chain so that they are joined as $\boldsymbol{d}{\in}\mathbb{R}^{n}$.
				This allows the \emph{determination of the clamping leg chain}. 
				The reason is that the contact force affects the platform through the links along the $i$-th kinematic chain.
				Since the passive revolute coupling joints only transmit forces, the forces' projection in platform coordinates has an \emph{intersection with the coupling joint}.
				Using the estimated external forces $\hat{\boldsymbol{F}}_\mathrm{ext}$ and the minimum distances $\boldsymbol{d}$ from (\ref{eq:MinDistd}), the second FNN categorizes the clamping contact into the $n$ classes $\{\mathrm{C}_1, \dots, \mathrm{C}_n\}$. The $i$-th class represents a clamping at the $i$-th kinematic chain.\\
		\subsection{Reaction Strategy: Structure Opening} \label{ssec:StructureOpening}
			If the second FNN classifies the $i$-th chain, a new target platform orientation $\tilde{\boldsymbol{x}}_\mathrm{r,d}$\footnote{Formulated for any spatial PR with three rotational platform coordinates} is demanded to open the clamping angle $q_{\mathrm{cl}}{=}\pi {-} q_{\mathrm{p}i}$ shown in Fig.~\ref{fig:reaktion_rueckz_strukoeff}(b).
			An orientation change
			\begin{align} \label{eq:retr_rot}
				[\alpha, \beta, \gamma]^\mathrm{T}=
				\mathrm{diag}\left(\mathrm{sgn} \left(\frac{\partial q_{\mathrm{cl}}} { \partial \boldsymbol{x}_\mathrm{r} } \right) \right) [|\alpha|, |\beta|, |\gamma|]^\mathrm{T}
			\end{align} 
			is formulated with the element-wise evaluation of the sign function over the partial derivatives. 
			The sign function is used to capture the \emph{direction of the steepest increase of the clamping angle} $q_{\mathrm{cl}}$ regarding the current orientation.
			In~(\ref{eq:retr_rot}), $|\alpha|, |\beta|$ and $|\gamma|$ are predefined rotational retraction angles.
			The structure opening is performed now with the Tait-Bryan angle residual of the difference rotation $\tilde{\boldsymbol{x}}_\mathrm{r,d}({^0\boldsymbol{R}_\mathrm{mP}^\mathrm{T}} {^0\boldsymbol{R}_\mathrm{d}})$ between the current orientation ${^0\boldsymbol{R}_\mathrm{mP}}$ and the desired orientation
			\begin{align} 
				{^0\boldsymbol{R}_\mathrm{d}}&={^0\boldsymbol{R}_\mathrm{mP}} {^\mathrm{mP}\boldsymbol{R}_\mathrm{d}}(\alpha,\beta, \gamma) \\
				\label{eq:RotDiff2}
				\text{with} \quad {^\mathrm{mP}\boldsymbol{R}_\mathrm{d}}(\alpha,\beta, \gamma)&= 
				\boldsymbol{R}_x(\alpha) \boldsymbol{R}_y(\beta) \boldsymbol{R}_z(\gamma).
			\end{align}
			The structure opening applied to the planar PR demonstrator simplifies to the scalar projection 
			\begin{align}
				\tilde{x}_\mathrm{r,d}=x_{\mathrm{r}}- \mathrm{sgn} 
				\left(\frac{\partial q_{\mathrm{p}i}} { \partial x_\mathrm{r} } \right) |\gamma|.
			\end{align}
		\subsection{Implementation} \label{ssec:Implementation}
			The complete reaction process consisting of the retraction movement and structure opening is termed \emph{reactive motion planner} and is shown in Algorithm \ref{alg:reactMotPl}.
			The inputs are the current joint angles $\boldsymbol{q}$, the platform pose $\boldsymbol{x}$, and the estimated forces $\hat{\boldsymbol{F}}_\mathrm{ext}$ by the MO in (\ref{eq:mo}). 	
			Important cases are an inaccurate estimate of the direction $\hat{\boldsymbol{n}}_\mathrm{f}$ of the LoA or misclassification of clamping. 
			\begin{figure}[b!]
				\vspace{1.5mm} 
				\centering
				\includegraphics[width=\columnwidth]{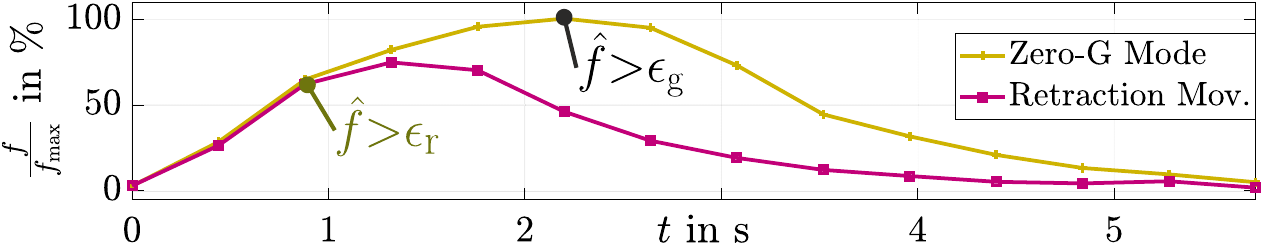}
				\caption{Normalized contact forces with a zero-g mode and retraction movement}
				\label{fig:Sicherheitsfunktionen_Grav_Reakt}
				\vspace{-1.5mm}
			\end{figure}	
			For this reason, a set of thresholds $\boldsymbol{\epsilon}_\mathrm{g}$ is defined. 
			As soon as $|\hat{F}_{\mathrm{ext},j}|{>}\epsilon_{\mathrm{g},j}$ is true, $\boldsymbol{F}_\mathrm{m}{=}\hat{\boldsymbol{g}}_x$ is~used to switch to zero-g mode in lines 2--3 assuming that the previously initiated reaction is disadvantageous.
			Thresholds for contact reaction are defined by $\boldsymbol{\epsilon}_\mathrm{r}{<}\boldsymbol{\epsilon}_\mathrm{g}$.
			If only $\epsilon_{\mathrm{r},j}$ is exceeded, the retraction movement or structure opening will be performed, as shown in Fig.~\ref{fig:Sicherheitsfunktionen_Grav_Reakt}.	
			Lines 5--8 form the retraction with lower impedances, while lines 9--16 provide the clamping leg classification. 
			The structural opening is implemented in line~17.
			In line 21, the reactions' results are summarized into $\tilde{\boldsymbol{x}}_\mathrm{d}{=}{[\tilde{\boldsymbol{x}}_\mathrm{t,d}^\mathrm{T}, \tilde{\boldsymbol{x}}_\mathrm{r,d}^\mathrm{T}]}^\mathrm{T}$ and a smooth reaction trajectory\footnote{Jerk-limited trajectory consisting of trapezoidal acceleration profiles} for the pose, velocity and acceleration of the platform coordinates forms the output.
			If no contact is detected, line 23 outputs with the continuation of a preplanned trajectory to complete a previous task. 
			\begin{figure}[tb!]
				\removelatexerror
				\vspace{1.5mm}
				\begin{algorithm}[H]
					\caption{Reactive motion planner for PRs}\label{alg:reactMotPl}
					{\small 
						\SetKwInOut{Input}{Input}
						\SetKwInOut{Output}{Output}
						\Input{$\hat{\mm{F}}_\ind{ext}, \mm{q},\mm{x}, \boldsymbol{\epsilon}_\mathrm{r}, \boldsymbol{\epsilon}_\mathrm{g}$}
						\Output{$\boldsymbol{x}_\mathrm{d}(t),\dot{\boldsymbol{x}}_\mathrm{d}(t),\ddot{\boldsymbol{x}}_\mathrm{d}(t)$}
						\uIf{$|\hat{F}_{\mathrm{ext},j}|{\ge}\epsilon_{\mathrm{g},j}$}
						{\tcp{Zero-g mode}
							Set $\boldsymbol{F}_\mathrm{m}{=}\hat{\boldsymbol{g}}_x$\;
							Set $\boldsymbol{x}_\mathrm{d}(t){=}\boldsymbol{x},\dot{\boldsymbol{x}}_\mathrm{d}(t){=}\boldsymbol{0},\ddot{\boldsymbol{x}}_\mathrm{d}(t){=}\boldsymbol{0}$\;
							
						}
						\uElseIf{$|\hat{F}_{\mathrm{ext},j}|{\ge}\epsilon_{\mathrm{r},j}$}
							{\tcp{Retraction movement}
							Set $\tilde{\boldsymbol{K}}_\mathrm{t,d}$ for more compliant robot behavior\;
							$\boldsymbol{r}_\mathrm{mP,LoA}\gets$ Minimal lever by (\ref{eq:CalcMPLoA})\;					
							$\hat{\boldsymbol{n}}_\mathrm{f}\gets$ Direction of line of action by (\ref{eq:CalcLoA})\;	
							$\tilde{\boldsymbol{x}}_\mathrm{t,d}\gets$ Calculate new platform end position by (\ref{eq:retr_tr})\;					
							$\boldsymbol{d} {\gets}\boldsymbol{0}$ Declare array for the minimal distances for $n$ chains\;
							\For{$i{=}1$ to $n$} 
							{\tcp{Calculate feature}
								$\boldsymbol{r}_{\mathrm{cJ}i}\gets$ $i$-th coupling joint position by serial forward kinematics\;		
								$d_i \gets$ Minimal distance $d_{\min,i}$ by (\ref{eq:MinDistd}) in row $i$ of $\boldsymbol{d}$\;
							}
							$b_\mathrm{clamp}\gets$ Binary output of 1st FNN for clamping classification\;
							\uIf{$b_\mathrm{clamp}$}
							{\tcp{Structure Opening}
								$i_\mathrm{cl}\gets$ Clamping chain classified by 2nd FNN\;
								$\tilde{\boldsymbol{x}}_\mathrm{r,d}\gets$ Calculate angle residual by (\ref{eq:retr_rot})--(\ref{eq:RotDiff2}) with joint angle from chain $i_\mathrm{cl}$\;
							}
							\Else 
							{$\tilde{\boldsymbol{x}}_\mathrm{r,d}\gets \boldsymbol{x}_\mathrm{r,d}$ Follow the preplanned orientation of the previous task\;
							}
							$\boldsymbol{x}_\mathrm{d}(t),\dot{\boldsymbol{x}}_\mathrm{d}(t),\ddot{\boldsymbol{x}}_\mathrm{d}(t) \gets$ Interpolation from current $\boldsymbol{x}_\mathrm{d}$ to $\tilde{\boldsymbol{x}}_\mathrm{d}{=}{[\tilde{\boldsymbol{x}}_\mathrm{t,d}^\mathrm{T}, \tilde{\boldsymbol{x}}_\mathrm{r,d}^\mathrm{T}]}^\mathrm{T}$\;
							}
						\Else 
						{\tcp{No reaction}
							$\boldsymbol{x}_\mathrm{d}(t),\dot{\boldsymbol{x}}_\mathrm{d}(t),\ddot{\boldsymbol{x}}_\mathrm{d}(t) \gets$ Follow the preplanned trajectory of the previous task\;
						}
					}
				\end{algorithm}
				\vspace{0mm}
			\end{figure}
	\section{Validation} \label{sec:validation}
		After the description of the experimental setup (\ref{ssec:Testbench}), the generalization of the clamping and chain classification algorithm is evaluated with experimental data (\ref{ssec:Classifier}). 
		This is followed by a comparison of the different reactions for collision and clamping (\ref{ssec:reaction_comparison}).
		Finally, the results of the different reaction strategies are evaluated under variation of platform velocities and contact stiffnesses (\ref{ssec:Retr_StructOp}).
		\subsection{Test Bench} \label{ssec:Testbench}
\begin{figure}[t!]
				\vspace{1.5mm} 
				\centering
				\includegraphics[width=\columnwidth]{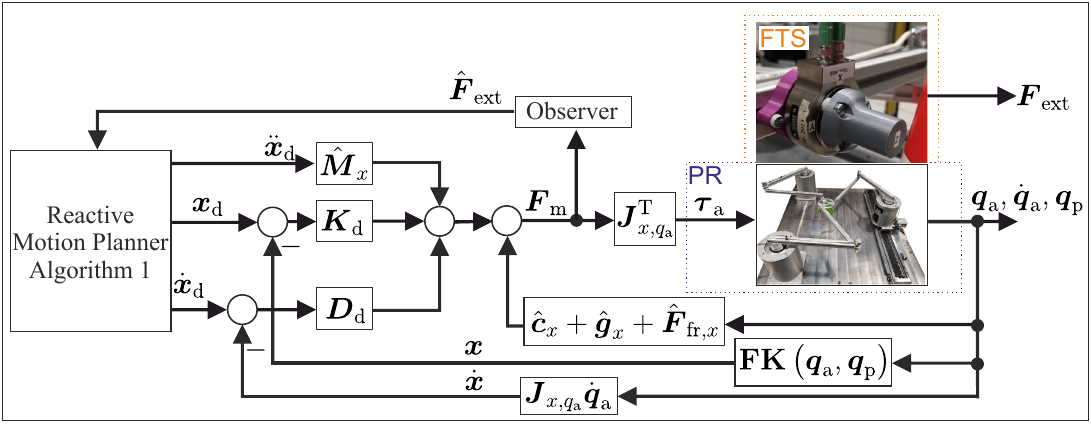}
				\caption{Block diagram with an extended experimental setup from~\cite{Mohammad.2023}}
				\label{fig:Ablauf}
				\vspace{-1.5mm}
			\end{figure}

			A force-torque sensor\footnote{KMS40 from Weiss Robotics} (FTS) is used to compare the different reaction strategies. 
			A ROS package\footnote{\url{https://github.com/ipa320/weiss_kms40}} of the FTS is integrated into the communication with the control system in \textsc{Matlab}/Simulink based on the EtherCAT protocol and a modification\footnote{\url{https://github.com/SchapplM/etherlab-examples}} of the open-source tool EtherLab\footnote{\url{https://www.etherlab.org}}. 
			Figure \ref{fig:Ablauf} represents the block diagram of the system with the reactive motion planner in Alg.~\ref{alg:reactMotPl} which is executed at a sampling rate of $\SI{1}{\kilo\hertz}$. 
			The Cartesian impedance control is parameterized with $\boldsymbol{K}_\mathrm{d}{=}\mathrm{diag}(\SI{2}{\newton/ \milli\meter}, \SI{2}{\newton/\milli\meter}, \SI{85}{Nm/\radian})$ and $D_{\xi,i}{=}1$. 
			The PR is torque-controlled since direct drives are used without gear friction. 
			The dynamics base parameters from~\cite{Thanh.2012} are used and symmetric leg chain parameters are assumed except for the friction. 
			The MO's gain is set to $k_{\mathrm{o},i}{=}\frac{1}{\SI{50}{\milli \second}}$.
			More information on the test bench is available in the authors' previous work~\cite{Mohammad.2023}.
		\subsection{Clamping Classification} \label{ssec:Classifier}
			In the following, the FNNs for clamping classification from Sec. \ref{ssec:ContactClass} are evaluated.	
			\begin{figure}[b!]
				\vspace{1.5mm} 
				\centering
				\includegraphics[width=\columnwidth]{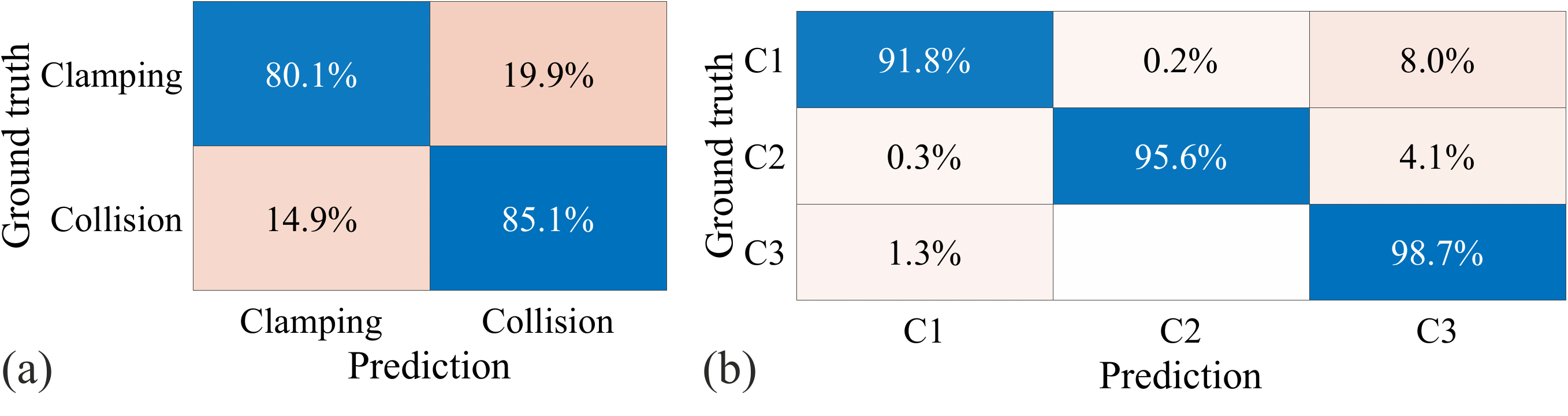}
				\caption{Row-normalized test results in confusion matrices (a) for the clamping (b) and chain (C) classification --- test and training are performed in different joint angle configurations of the PR}
				\label{fig:class_clamping_chain}
				\vspace{-1.5mm}
			\end{figure}	
			Fig.~\ref{fig:class_clamping_chain}(a) depicts the row-normalized test results of the classification from a detected contact into a collision and clamping. 
			Both FNNs are trained and evaluated with known inputs ($\hat{\boldsymbol{F}}_\mathrm{ext},\boldsymbol{d}$) and labeled outputs from clamping at the joints and collisions across the structure of the PR in different joint angle configurations.
			In total, measurements of three robot configurations are available, each with three clamping cases and seven collisions (six links and platform). 
			The data set consists of 80k measurements samples with $|\hat{F}_{\mathrm{ext},j}|{>}\epsilon_{\mathrm{r},j}$, which is split into $70\%,30\%$ for training and testing.
			To evaluate the generalization of the FNN, training and testing are performed with data from different configurations of the PR. 			
			The training dataset is balanced and used to optimize the hyperparameters (network architecture and regularization term).
			The results of grid search hyperparameter optimization are two FNNs with five hidden layers and 30 (25) neurons per hidden layer for classification of clamping (of the affected leg chain).
			$80\%$ and $85\%$ of all clamping and collision test cases are correctly classified.
			The effects of misclassification are limited by the thresholds in $\boldsymbol{\epsilon}_\mathrm{g}$, so that the worst-case response is a gravity-compensated PR.
			
			Figure~\ref{fig:class_clamping_chain}(b) shows the test results of clamping chain classification. 
			Here, a clamping chain is correctly categorized with an accuracy of more than $90\%$, which can be attributed to the use of the minimum distances $\boldsymbol{d}$ as a physically modeled feature in addition to the estimated external forces in $\hat{\boldsymbol{F}}_\mathrm{ext}$.
			Since the inputs to the FNNs are computed at the sampling rate, the classifications are executable at the same time step as the detection to enable an immediate reaction. 
			
		\subsection{Comparison of Different Reaction Methods} \label{ssec:reaction_comparison}
					\begin{figure}[b!]
			\vspace{1.5mm} 
			\centering
			\includegraphics[width=\columnwidth]{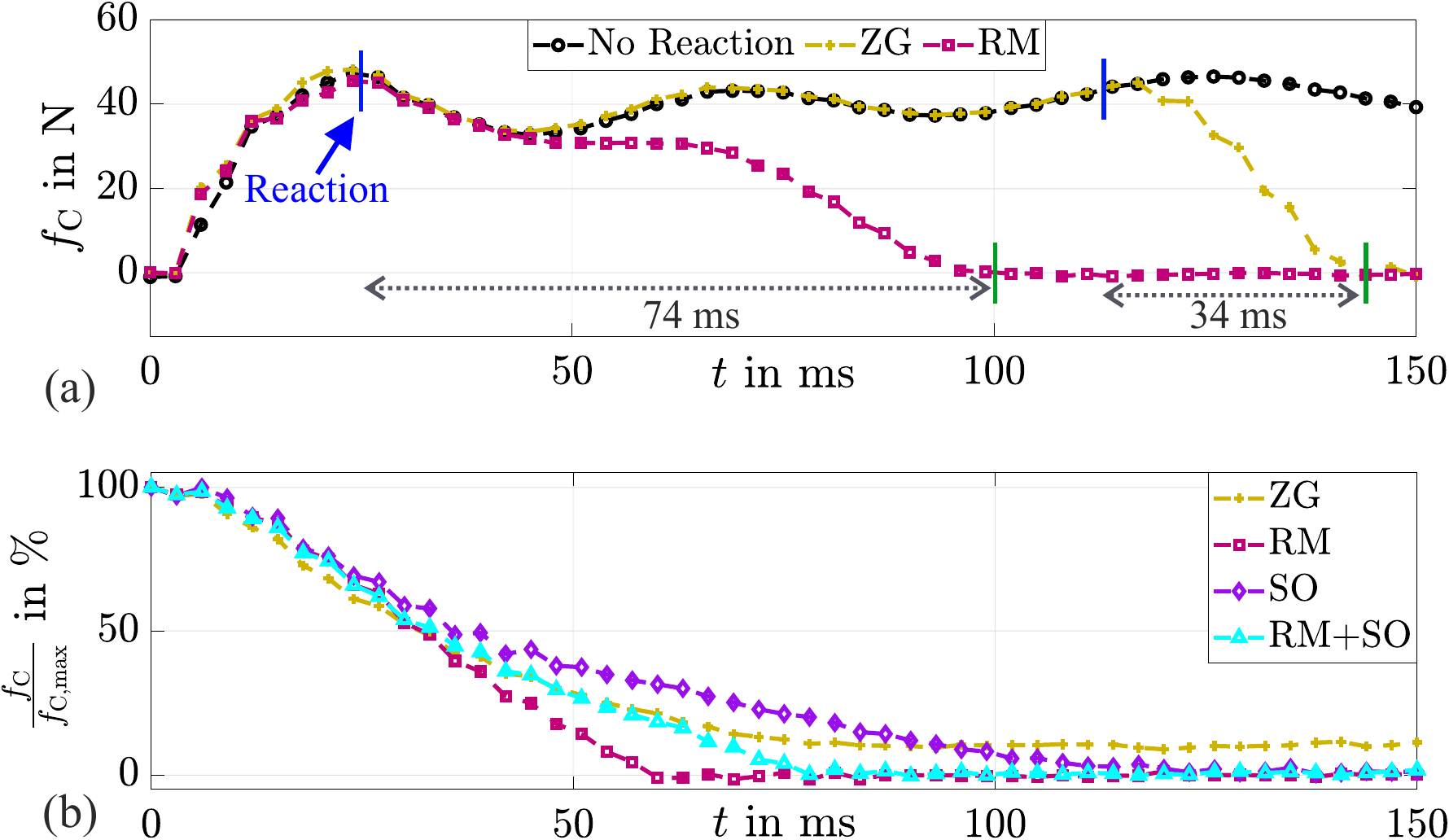}
			\caption{Compared reactions (a) to collisions (b) and clamping. \textcolor{blau}{Detection and start of the reaction} are shown by the blue marker. \textcolor{gruenmarker}{Contact termination} is depicted by the green marker}
			\label{fig:coll_comp}
			\vspace{-1.5mm}
		\end{figure}
			A comparison of zero-g mode (ZG), retraction movement (RM), structure opening (SO), and a combination of the latter two (RM+SO) for collisions and clamping follows.
			With the conducted real-world experiments on the planar PR, the effectiveness of the reactive motion planner in Alg.~\ref{alg:reactMotPl} is demonstrated.
			The FTS is used only to measure contact forces $f_\mathrm{C}$ for evaluation.								
			The contact reaction thresholds are $\boldsymbol{\epsilon}_\mathrm{r}^\mathrm{T}{=} [\SI{10}{\newton},\SI{10}{\newton}, \SI{1}{Nm}]$ and for a switch into zero-g mode $\boldsymbol{\epsilon}_\mathrm{g}{=}4\boldsymbol{\epsilon}_\mathrm{r}$.			
			The distance of retraction and the orientation change of the structure opening are chosen as $d_\mathrm{react}{=}\SI{50}{\milli \meter}$ and $|\gamma|{=}\SI{5}{\degree}$.			
			All reactions are determined from modeled quantities and proprioceptive information. 
			The data is synchronized in advance to the respective contact beginning for the comparison of the results.					

			\subsubsection{Collision}
\begin{figure}[b!]
					\vspace{1.5mm} 
					\centering
					\includegraphics[width=\columnwidth]{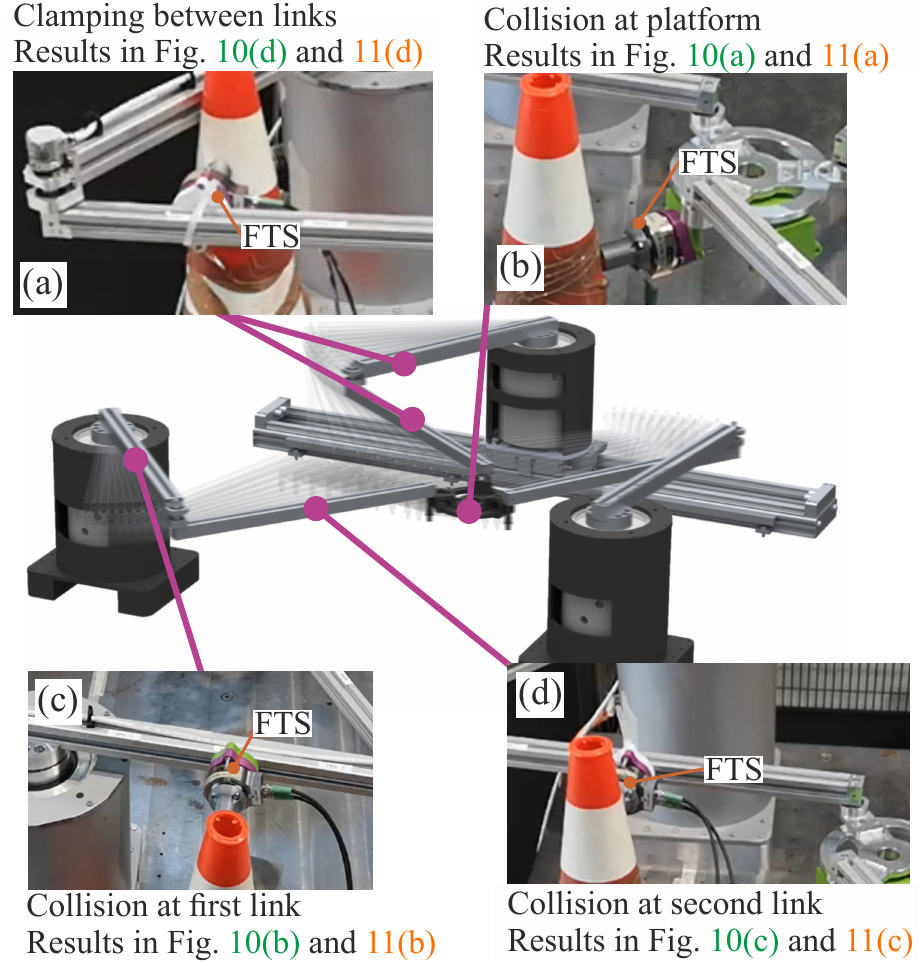}
					\caption{(a) Clamping at a chain and collisions at the (b) platform and (c,d) links to test contact reactions for variations of \textcolor{gruen}{impedance-related stiffness} and \textcolor{orange}{platform velocities}}
					\label{fig:CollDet_ablauf}
					\vspace{-1.5mm}
				\end{figure}

				Figure~\ref{fig:coll_comp}(a) depicts the repeated reaction results for platform collisions with a fixed obstacle like in Fig.~\ref{fig:CollDet_ablauf}(b) without a reaction, with ZG and RM.
				The collision with the maximum force $f_\mathrm{max}{=}\SI{50}{\newton}$ is selected to show the utility of the reaction thresholds $\boldsymbol{\epsilon}_\mathrm{r}$ and $\boldsymbol{\epsilon}_\mathrm{g}$ for a correctly and falsely classified contact.
				The platform velocity and reaction stiffness during the collision are $||\dot{\boldsymbol{x}}_\mathrm{t}||_2{=}\SI{0.25}{\meter / \second}, \tilde{K}_{\mathrm{t}i,\mathrm{d}} {=} \SI{2}{\newton / \milli \meter}$.
				The RM occurs before the ZG due to $\boldsymbol{\epsilon}_{\mathrm{r}}{<}\boldsymbol{\epsilon}_{\mathrm{g}}$. 			
				The ZG shows a shorter duration ($\SI{34}{\milli \second}$) from the initiation of the reaction (blue markers) to contact termination (green markers and indicated by $f_\mathrm{C}{=}\SI{0}{\newton}$) than the RM ($\SI{74}{\milli \second}$).
				A reason is the initiation of the reaction trajectory from the target pose $\boldsymbol{x}_\mathrm{d}$ (see line 21 in Alg. \ref{alg:reactMotPl}).
				In the contact case, $\boldsymbol{x}_\mathrm{d}$ is in the interior of the contact object. 
				The smooth retraction trajectory is planned from $\boldsymbol{x}_\mathrm{d}$ to $\tilde{\boldsymbol{x}}_\mathrm{d}$ causing the trajectory to begin in the interior of the contact object.
				Alternatively, a start from the actual pose $\boldsymbol{x}$ at the contact time is possible to obtain a faster but non-smooth reaction trajectory of the PR.								
			\subsubsection{Clamping}
				Figure~\ref{fig:coll_comp}(b) presents the results of clamping experiments performed with the same velocity and stiffness conditions ($||\dot{\boldsymbol{x}}_\mathrm{t}||_2{=}\SI{0.05}{\meter / \second}, \tilde{K}_{\mathrm{t}i,\mathrm{d}} {=} \SI{2}{\newton / \milli \meter}$).
				Shown are the forces $f_\mathrm{C}$ normalized to the respective maximum $f_\mathrm{C,max}$ after the initiation of the reaction with a ZG, RM, SO and the combination RM+SO. 
				It is noticeable that the reaction ZG does not eliminate the contact. 
				This can be attributed to non-compensated dynamic effects such as cogging torques in the drives.
				The reactions with RM terminate the contact the fastest.
				The combined strategy RM+SO indicates a slower contact removal compared to RM, suggesting that the effect of SO on reaction is unfavorable regarding contact removal. 
				However, the results show in all cases a contact termination in less than $\SI{110}{\milli \second}$.	
		\subsection{Stiffness and Velocity Variation} \label{ssec:Retr_StructOp}	

			Figure~\ref{fig:CollDet_ablauf}(a)--(d) shows the process for measuring the contact forces during clamping at a joint, collisions at the mobile platform, and the two links of a kinematic chain.
			For each of the four contact situations (three collisions, one clamping), the reaction strategies RM for collisions and SO for clamping are evaluated at different platform velocities $\SI{0.05}{\meter / \second}{<}||\dot{\boldsymbol{x}}_\mathrm{t}||_2{<}\SI{0.42}{\meter / \second}$ and reaction stiffnesses $\SI{0.1}{\newton / \milli \meter} {\le} \tilde{K}_{\mathrm{t}i,\mathrm{d}} {\le} \SI{2}{\newton / \milli \meter}$.
			In each experiment, the stiffness of the Cartesian impedance control is initially set to $K_{\mathrm{t}i,\mathrm{d}} {=} \SI{2}{\newton / \milli \meter}$.
			\subsubsection{Stiffness Variation}
\begin{figure}[t!]
					\vspace{1.5mm} 
					\centering
					\includegraphics[width=\columnwidth]{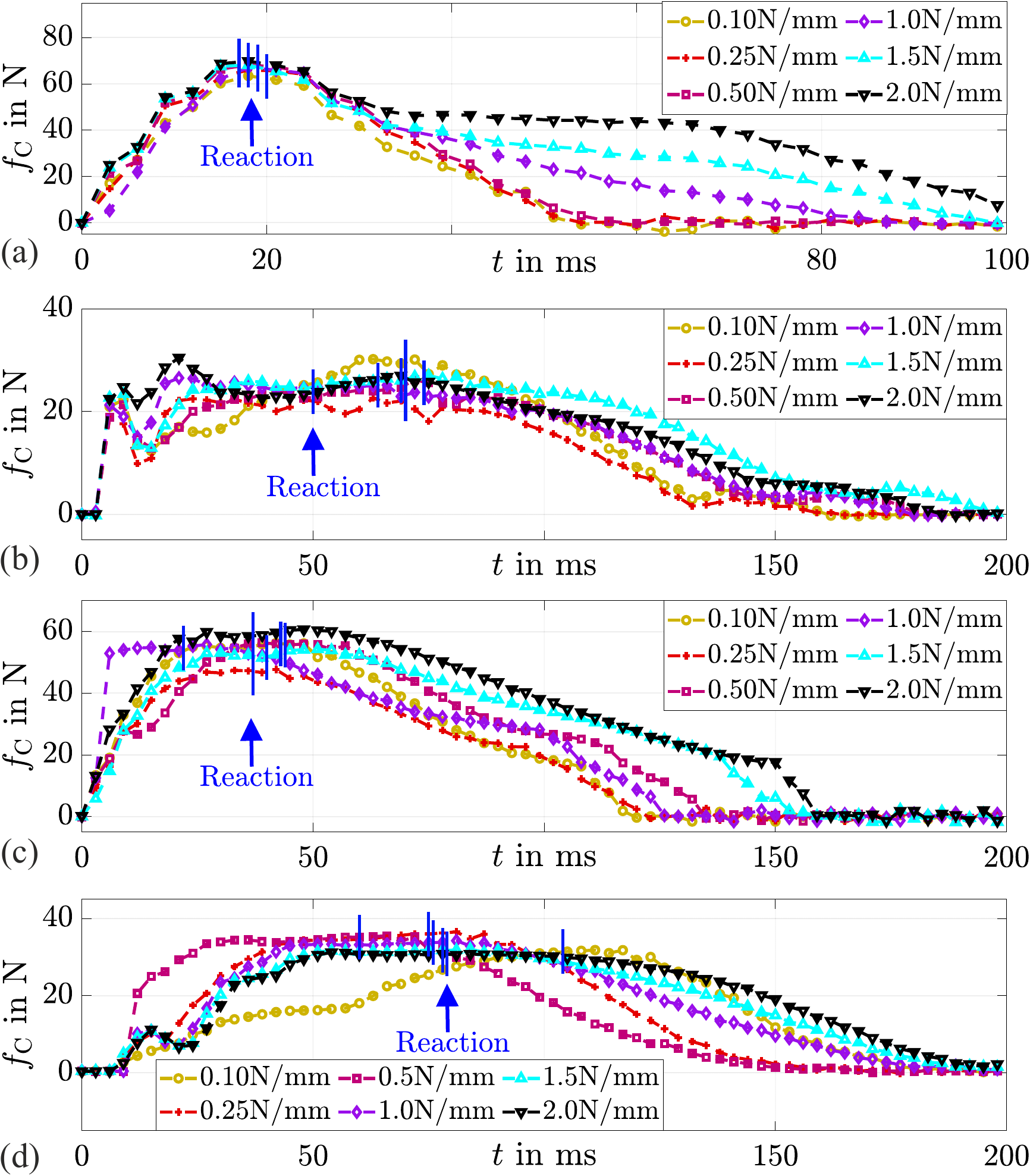}
					\caption{Stiffness variation for reactions on contacts (a) at the mobile platform, (b)--(c) at the first and second link, (d) and during clamping. \textcolor{blue}{Detection and start of the reaction} are shown by the blue marker}
					\label{fig:StiffVar}
					\vspace{-1.5mm}
				\end{figure}

				Figure~\ref{fig:StiffVar} depicts the results of the different \emph{reaction} stiffnesses at the same platform velocity $||\dot{\boldsymbol{x}}_\mathrm{t}||_2{\approx}\SI{0.4}{\meter / \second}$.
				It appears from the platform collisions in Fig.~\ref{fig:StiffVar}(a) that the retraction movement limits the maximum force to $\SI{70}{\newton}$. 
				From the reaction time steps at $15$--$\SI{18}{\milli \second}$, the curves differ depending on the reaction stiffness. 
				For the most compliant setting, contact is terminated at $t{=}\SI{55}{\milli \second}$, while for the stiffest mode, contact disappears from $t{=}\SI{100}{\milli \second}$ on.	
				This observation also holds for the results of collisions at the second link in Fig.~\ref{fig:StiffVar}(c) and clamping at a joint in Fig.~\ref{fig:StiffVar}(d).
				A faster decaying contact force occurs with a more compliant control.
				This is least apparent for collisions at the first link in Fig.~\ref{fig:StiffVar}(b). 
			
				The underlying reason is the smooth reaction trajectory starting in the interior of the contact object. 
				As long as this holds true, a stiffer controller will result in higher contact forces by the robot in the post-contact phase. 
			\subsubsection{Velocity Variation}
\begin{figure}[t!]
					\vspace{2mm} 
					\centering
					\includegraphics[width=\columnwidth]{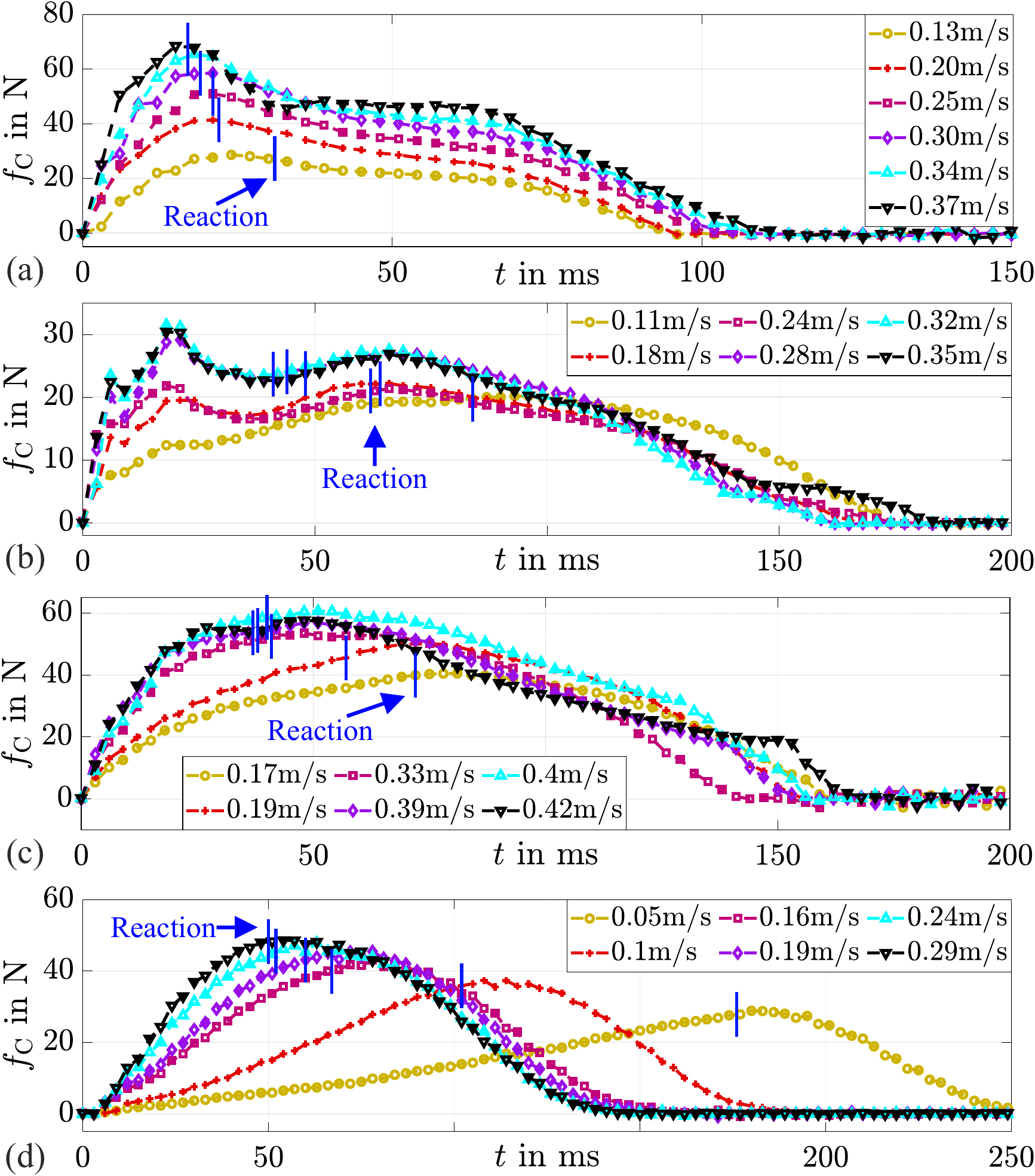}
					\caption{Velocity variation for reactions on contacts (a) at the mobile platform, (b)--(c) at the first and second link, (d) and during clamping. \textcolor{blau}{Detection and start of the reaction} are shown by the blue marker}
					\label{fig:VelocVar}
					\vspace{-1.5mm}
				\end{figure}

				Figure~\ref{fig:VelocVar} presents the results of the retraction movement on the mobile platform and links, as well as the opening of a clamping. 
				Platform contacts with velocities of $0.13$--$\SI{0.37}{\meter / \second}$ are detected within $\SI{40}{\milli \second}$, after which the retraction movement terminates the contact within $\SI{90}{\milli \second}$.
				For collisions at the second link and clamping, it can be noted from Fig.~\ref{fig:VelocVar}(c,d) that a maximum contact force of $\SI{60}{\newton}$ occurs as a result of the reaction. 
				The lowest occurring contact forces $\SI{30}{\newton}$ are shown at the collision results of a first link in Fig.~\ref{fig:VelocVar}(b). 
				The reason is the single influence of a drive on the first link of its kinematic chain, while contacts on the other bodies of the PR involve several drives.\\
				Finally, for the collision and clamping experiments, the reaction strategies based on proprioceptive information are found to terminate contact within a maximum of $\SI{130}{\milli \second}$. 
	\section{Conclusion} \label{sec:conlusions}
		The contribution of this work consists of \emph{reaction strategies} based on \emph{proprioceptive information} in \emph{collision} and \emph{clamping} scenarios with \emph{parallel robots} (PRs). 
		The direction of the model-based estimated line of action allows immediate retraction using the translational coordinates without knowing the exact contact location.
		Employing this reaction, forces during collisions on the mobile platform and the links of a kinematic chain are limited to a maximum of $\SI{70}{\newton}$.
		Clamping hazards at the joints of a kinematic chain are classified by two feedforward neural networks with an accuracy of $80\%$ and assigned to the clamping chain with $90\%$. 
		A gradient-based projection of the clamping angle onto the platform orientation results in a structural opening.
		Both reaction strategies are extended by a stiffness reaction so that the PR becomes more compliant in the contact case.
		The results show that in all cases the PR terminates the contacts in less than $\SI{130}{\milli \second}$. 
		The retraction movement performed best since it requires no determination of the clamping leg chain and terminates the contact most rapidly.
		Switching to zero-g mode showed a $\SI{40}{\milli \second}$ shorter contact duration than retraction in the collision case, since the smooth reaction trajectory is planned from the desired platform pose.
		Furthermore, compared to zero-g mode, the presented methods were able to completely remove the clamping contact.
		Future research focuses on the application of the presented methods to spatial PRs, investigating reaction strategies to a proprioceptively isolated collision and the improvement of trajectory planning based on the actual platform pose at contact time.
	\addtolength{\textheight}{0cm} 
	
	
	\section*{Acknowledgment}
	The authors acknowledge the support by the German Research Foundation (DFG) under grant number 444769341.
	\bibliographystyle{IEEEtran}
	\bibliography{literatur}	
\end{document}